**Title:**

Adapting and evaluating a deep learning language model for clinical why-question answering

**Authors:**

[1]Andrew Wen

[3,4]Mohamed Y. Elwazir

[1]Sungrim Moon

[1,2]Jungwei Fan*

[1]Department of Health Sciences Research

[2]Robert D. and Patricia E. Kern Center for the Science of Health Care Delivery

[3]Department of Cardiovascular Medicine

Mayo Clinic, Rochester, United States

[4]Department of Cardiology, Faculty of Medicine

Suez Canal University, Ismailia, Egypt

*Corresponding author: fan.jung-wei@mayo.edu

200 1st Street SW, RO-HA-2-CSHCD, Rochester, MN 55905

(507) 538-1191

**ABSTRACT:**

**Objectives**

To adapt and evaluate a deep learning language model for answering why-questions based on patient-specific clinical text.

**Materials and Methods**

Bidirectional encoder representations from transformers (BERT) models were trained with varying data sources to perform SQuAD 2.0 style why-question answering (why-QA) on clinical notes. The evaluation focused on: 1) comparing the merits from different training data, 2) error analysis.

**Results**

The best model achieved an accuracy of 0.707 (or 0.760 by partial match). Training toward customization for the clinical language helped increase 6% in accuracy.

**Discussion**

The error analysis suggested that the model did not really perform deep reasoning and that clinical why-QA might warrant more sophisticated solutions.

**Conclusion**

The BERT model achieved moderate accuracy in clinical why-QA and should benefit from the rapidly evolving technology. Despite the identified limitations, it could serve as a competent proxy for question-driven clinical information extraction.

# INTRODUCTION

The reasoning and decision-making in clinical practice can be naturally framed as a series of questions and answers. Automated question-answering (QA) has long been considered a feat in artificial intelligence (AI) and is vitally researched for clinical applications. Among the diverse information needs, why-QA is a distinct category that deals with cause, motivation, circumstance, and justification. In terms of prevalence, 20% of the top ten question types asked by family physicians[1] can actually be paraphrased into a why-question. Clinical why-QA is important because: 1) the ultimate task resembles expert-level explanatory synthesis of knowledge and evidence, 2) it would enable identifying reasons for the decisions documented in clinical text.

The current study concentrates on the second scenario above, a modest yet very useful task of reason identification. Essentially, the system has to identify the literal reason regarding certain decision specific to a patient, e.g. why was his dobutamine stress test rescheduled? → "hypotension" (from note text). In non-medical domains, the counterpart to such document-based QA is known as reading comprehension QA (RCQA), with competitive open challenges and richly-annotated corpora. SQuAD 2.0[2] is an iconic RCQA corpus and challenge, which features the requirement for a system to refrain from answering when there is no suitable answer present in the text. A language model that has caught wide attention was the bidirectional encoder representations from transformers (BERT)[3] and its evolving derivatives[4], for their high performance not only in SQuAD 2.0 but in multiple natural language understanding challenges.

As an initial step toward developing a clinical reason identification system, this study adapted the BERT model for clinical why-QA. We found domain customization was critical to performance, with a best achieved accuracy of 0.707 (or 0.760 by partial match). More importantly, our error analysis helped understand the data issues, the system behavior, and areas to improve on.

## BACKGROUND

In the following we introduce several existing resources that are important to our methods.

## SQuAD

The Stanford Question Answering Dataset (SQuAD)[5] was created to promote RCQA research and application development. We followed the SQuAD 2.0 task setting, because it can be critical to have the system refrain from making false suggestions especially in some clinical applications. **Figure 1** illustrates a typical training instance in the SQuAD 2.0 format, consistent with that used in our experiments.

>> Figure 1

## emrQA

The emrQA[6] is a large training set annotated for RCQA in the clinical domain. It was generated by template-based semantic extraction from the i2b2 NLP challenge datasets[7]. The current emrQA release includes more than 400,000 QA pairs, of which 7.5% involve a why-question.

**BERT**

BERT[3] represents a state-of-the-art language model that leverages deep bidirectional self-attention learning. The pre-training phase of BERT learns a transferrable representation, which can be followed by a fine-tuning phase where the actual task-specific (e.g., RCQA) training takes place. Due to the immense memory demand for training $BERT_{large}$, we used $BERT_{base}$ for our experiments without losing conceptual generality.

**Clinical BERT**

Alsentzer et al. used approximately 2 million clinical notes from the MIMIC-III v1.4 database[8] and pre-trained a Clinical BERT model[9]. They made it publicly available; otherwise it originally took about 17 days of computational runtime by a single GeForce GTX TITAN X 12 GB GPU.

**MATERIAL AND METHODS**

**Preparation of the training data**

emrQA<sub>why</sub>

This was our core training data, by selecting emrQA entries with a why-question. Additional processes included: 1) removing the "heart-disease-risk" subset due to problematic index in the "evidence_start" field, 2) retaining QAs where the question had one and only one answer, in conformity with SQuAD 2.0 setting, 3) merging in a small set of our previously annotated clinical why-QAs[10], and 4) programmatically constructing a set of unanswerable QAs where neither the question's key concept nor the answer was present in the note text. We obtained 27,762 answerable QAs and 2,839 unanswerable QAs, all formatted like **Figure 1**. Lastly, the data was split into train/dev[elopement]/test partitions with 250/90/250 disjoint clinical notes, corresponding to 12,741/4,315/13,545 QAs. The dev partition was set aside to learn the optimal cutoff threshold for the system to refrain from answering.

i2b2notes<sub>pre</sub>

A set of 1,474 i2b2 notes with 106,952 pre-chunked sentences was available to us and amenable to BERT pre-training. Given the manageable size, we undertook this pre-training and evaluated its usefulness for domain customization in comparison to the more heavily trained Clinical BERT.

SQuAD<sub>why</sub>

A phenomenon commonly observed in biomedical NLP is the smoothing effect introduced by inclusion of off-domain training data. To experiment with this aspect, we extracted 1,833 why-

QAs from SQuAD 2.0 (hence SQuAD$_{why}$). SQuAD$_{why}$ was run as a "pre"-fine-tuning step to prime BERT into the why-QA genre.

**Training and tuning of the QA models**

To assess the benefits of different data sources, we experimented with five models trained by incremental levels of domain and task customization. **Figure 2** illustrates the paths of configuring these models. BERT$_{base}$ was the original general-purpose model. The emrQA$_{why}$ served as the core task-specific training set and (when used alone) as the baseline for benchmarking other enhancements. Clinical BERT and i2b2Notes$_{pre}$ represented domain adaptation; the former used a much larger corpus and more diverse note types. The SQuAD$_{why}$ (1,833 general English why-QAs) was an optional fine-tuning step to assess how an off-domain training set might benefit the model. Each fine-tuning experiment was run with 5 epochs, batch_train_size=32, learning_rate=3e-5, and max_seq_length=128. The jobs were run on a Tesla V100 with compute capability 7.0 and 18 GB of memory.

>> Figure 2

**Evaluation and error analysis**

The evaluation was based on the standard SQuAD 2.0 metrics, comparing the system answer to the gold by token-wise exact and partial matches. Each partial match was weighted by using f1-measure between the predicted and the gold bags of tokens[11]. We computed the accuracies and included a break-down summary of the answerable versus unanswerable QAs. After the optimal configuration emerged, we doubled the epochs to 10 and trained a separate model for the final precision-recall and error analysis. The error analysis focused on false negatives (FNs), by randomly sampling 100 QAs where the system answer had completely no overlap tokens (including those refrained) with the gold answer. A cardiologist (MYE) manually reviewed the set and recorded his assessment.

## RESULTS

### Accuracy of the why-QA models

The performance metrics of the differently trained models are summarized in **Table 1**. In general we can see the refraining mechanism worked well (the NoAns column). The "pre"-fine-tuning by $SQuAD_{why}$ appeared beneficial (almost 3% accuracy increase from the baseline), suggesting that BERT learned certain genre characteristics even from a non-medical corpus. Pre-training using the 1,474 notes of $i2b2notes_{pre}$ lifted the accuracy up about 3%, but could not beat the 6% boost by the lavishly trained $ClinBERT_{pre}$ using 2 million notes. In the end we combined the best configurations into training a single model, which achieved an accuracy of 0.700 (or 0.757 with partial match). The extended training with 10 epochs resulted in a marginal increase in accuracy.

**Figure 3** shows the precision-recall tradeoff of the final 10-epoch model. Overall the system appeared to conservatively favor higher precision versus recall, while maintaining the precision above or around 0.8 until the upper-bound recall due to the refraining. As for the cost of time, the best configuration (fine-tuned by $SQuAD_{why}$ then $emrQA_{why}$, on top of $ClinBERT_{pre}$) took 53 minutes to train with 5 epochs and 64 minutes with 10 epochs.

**Table 1.** Accuracy of differently trained models on the test set

| Model | Full test set: 13,545 QAs | | Test HasAns: 12,376 QAs | | Test NoAns: 1,169 QAs | |
|---|---|---|---|---|---|---|
| | Exact | Partial | Exact | Partial | Exact | Partial |
| BERT$_{base}$ + emrQA$_{why}$ | 0.633 | 0.688 | 0.599 | 0.659 | 0.995 | 0.995 |
| BERT$_{base}$ + SQuAD$_{why}$ + emrQA$_{why}$ | 0.660 | 0.728 | 0.628 | 0.703 | 0.994 | 0.994 |
| BERT$_{base}$ + i2b2Notes$_{pre}$ + emrQA$_{why}$ | 0.663 | 0.718 | 0.631 | 0.692 | 0.997 | 0.997 |
| BERT$_{base}$ + ClinBERT$_{pre}$ + emrQA$_{why}$ | 0.695 | 0.744 | 0.666 | 0.720 | 0.994 | 0.994 |
| BERT$_{base}$ + ClinBERT$_{pre}$ + SQuAD$_{why}$ + emrQA$_{why}$ | **0.700** | **0.757** | 0.672 | 0.735 | 0.995 | 0.995 |
| BERT$_{base}$ + ClinBERT$_{pre}$ + SQuAD$_{why}$ + emrQA$_{why}$ (10 epochs) | 0.707 | 0.760 | 0.679 | 0.737 | 0.999 | 0.999 |

*HasAns: answerable according to the gold, NoAns: unanswerable according to the gold.

>> Figure 3

## Error analysis with a focus on the false negatives

Given that the system performed decently on the NoAns QAs, we paid attention to those where the gold indeed had an answer. Based on review by the physician (MYE), we categorized the 100 FNs missed by our final best model as in **Table 2** and elaborated below.

Unanswerable

*a) Vague question (6%)*

The question did not make sense, likely due to the fact that emrQA synthetically derived the questions from i2b2 NLP annotations. For example, "why did the patient have removal?"

*b) Expert deemed unanswerable using only text (8%)*

There was no clear trace of reasoning mentioned in the text to support the answer without preexisting medical knowledge, or the correct answer was not even present.

System answered

*c) Expert judged the system as acceptable as the gold (6%)*

There were two scenarios here. The first was that the system picked a conceptually synonymous answer, e.g., "bacteremia" versus "sepsis". The second revealed incompleteness in some gold annotations. For example, the reasons why one patient was on nitroglycerin actually included both "shortness of breath" and "chest pain", but the gold had only the former.

*d) Expert sided with the system against the gold (12%)*

The domain expert considered the system's answer to be more suitable than the gold. For example, when asked to identify the reason behind a Flagyl prescription, the system picked "aspiration pneumonia" instead of the "elevated white count" by the gold annotation.

*e) Real FN (18%)*

The system did not appear to really understand the nuances such as the indication versus the target effect. For example, in one case "diuresis" was picked as the reason for Lasix drip, while the gold had "decreased urine output". In the other example, the mentioned side effect of Celebrex was mistaken as the reason for prescription, seemingly because they co-occurred within the same sentence.

*f) Expert disagreed with both the system and gold (7%)*

The physician judged that neither the system nor the gold picked the correct answer.

System refrained

*g) Real FN (24%)*

The system not only refrained from answering but the lower-ranked candidate answers did not appear to be viable either.

*h) Correct answer ranked second place (19%)*

As a rescue investigation, we found that in 19% of the FNs the system actually had the correct answer but ranked it secondary to the refraining decision.

**Table 2.** Error analysis of 100 FNs by the best model

| Main category | Subcategory | Count | Subtotal |
|---|---|---|---|
| Unanswerable | a) Vague question | 6 | 14 |
| | b) Expert deemed unanswerable using only text | 8 | |
| System answered | c) Expert judged the system acceptable as the gold | 6 | 18 |
| | d) Expert sided with the system against the gold | 12 | |
| | e) Real FN | 18 | 68 |
| | f) Expert disagreed with both the system and gold | 7 | |
| System refrained | g) Real FN | 24 | |
| | h) Correct answer ranked second place | 19 | |

*Note that the rightmost column stands for a "redemption" perspective: 14% that the system was not supposed to make it, 18% where the system answer was actually right, and 68% that the system was truly attributed for the FN.

## DISCUSSION

Our incremental training source comparisons proved to be informative. The 3% accuracy increase by the SQuAD$_{why}$ pre-tuning suggested that a close-genre corpus, even off-domain, could benefit the end task. In alignment with intuition, Clinical BERT as an extensively pre-trained in-domain model was a vital booster to accuracy (6% improvement). Noteworthy on the other hand, the 6% was earned by a hefty 2 million training notes. Given the auxiliary finding that even the 1,474 notes of i2b2notes$_{pre}$ made a 3% increase, it raised the question whether the benefit could have been saturated much earlier before the training data was increased into the millions.

The results revealed much room for improvement both in terms of the data preparation and model optimization. Even though erring conservatively is desirable for many precision-oriented applications, the over-refraining tendency implied that our heuristically constructed unanswerable instances were noisy. Besides, the error analysis discovered various issues (e.g., nonsense questions and inaccurate answers) in emrQA, which was efficient for producing massive training data but regrettably could not match the quality of manually-authored QAs like the SQuAD corpus.

Despite the rapidly evolving field, our study exposed issues and challenges that are general to deep learning language models as applied in clinical why-QA. Some off-the-mark answers by the system showed that BERT might have just leveraged adjacent cues or recurring associations, instead of true understanding. In that same vein, those deemed by the physician as unanswerable by the text indicated that why-QA probably should not be framed simply as an RCQA task, i.e., much richer contexts (including external knowledge sources) are needed both at the training and the reasoning phases. Furthermore, the long documents and existence of multiple viable answers distinguish clinical why-QA as a unique challenge that warrants redesign in the annotation and evaluation approaches.

**CONCLUSION**

The BERT language model was evaluated for the task of clinical why-QA. The best accuracy was 0.707 (or 0.760 with partial match), specifically benefiting from domain- and genre-customization. The error analysis indicated improvements to be needed in the training data preparation and even redesign of the fundamental task. Although at its current state the model is premature for truly intelligent why-reasoning, we propose to use it practically as a question-driven clinical information extraction tool for detecting reasons with explicit cues in text.

**ACKNOWLEDGEMENT**

De-identified clinical records used in this research were provided by the i2b2 National Center for Biomedical Computing funded by U54LM008748 and were originally prepared for the Shared Tasks for Challenges in NLP for Clinical Data organized by Dr. Ozlem Uzuner, i2b2 and SUNY.

**COMPETING INTERESTS**

The authors claim no competing interests as associated with the research.

**FUNDING**

The study was supported by the Robert D. and Patricia E. Kern Center for the Science of Health Care Delivery, Mayo Clinic.

```
{
    "note_id": "doc_12345"
    "note_text": "This is a mock sentence. This is a mock sentence.
 This is a mock sentence… The patient's potassium has been low,
 which requires regular monitoring.  This is probably due to the
 extensive course of gentamicin… This is a mock sentence."
    "qas": [
        {
            "qa_id": "qa_0000001",
            "question": "Why the patient's potassium was low?",
            "answerable": true,
            "answer": [
                "text": "gentamicin",
                "begin_offset": 678
            ]
        }
        …
    ]
}
```

**Figure 1.** Illustration of a SQuAD-style training instance for clinical why-QA.

*The format conforms to JSON syntax, where the gray highlight here is for helping readers locate the answer. For readability, some of the field names have been modified and are different from that in the original SQuAD 2.0.

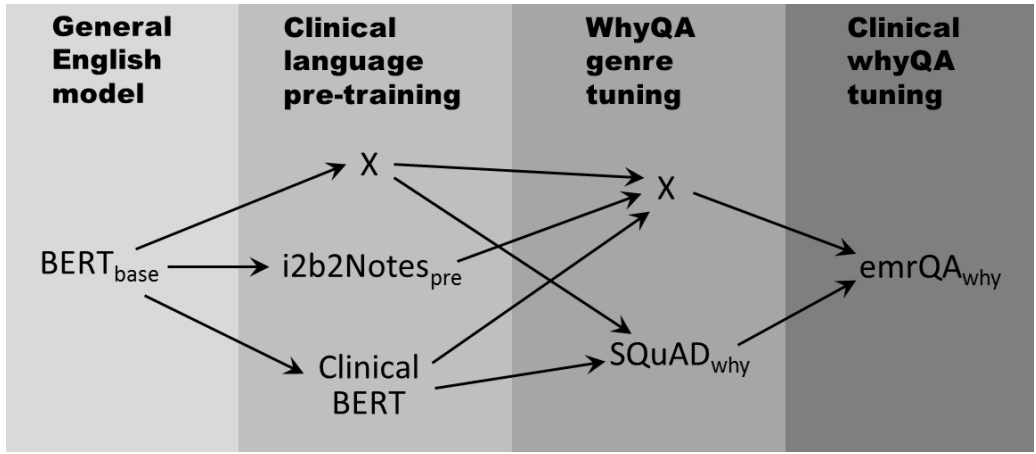

**Figure 2**. Different learning configurations to explore the effect of domain-, genre- or task-specific customization.

*Each "X" means no enhancement performed. The grayscale roughly corresponds to the level of customization by the training data, from general (lighter) to specific (darker).

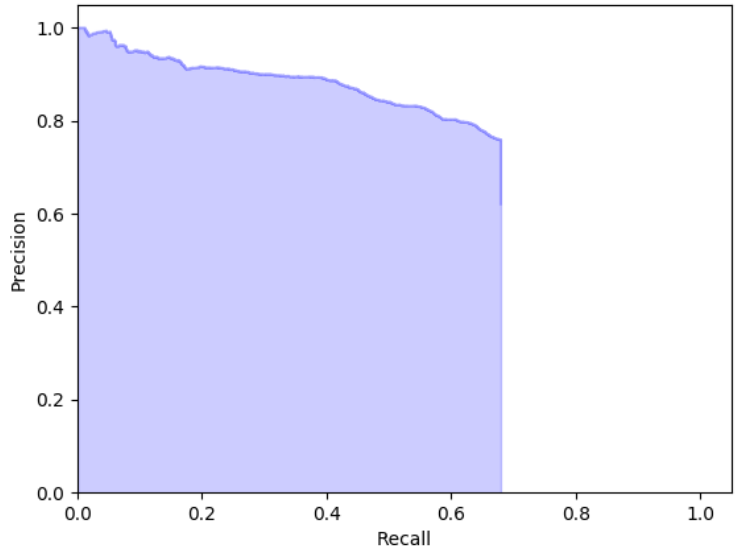

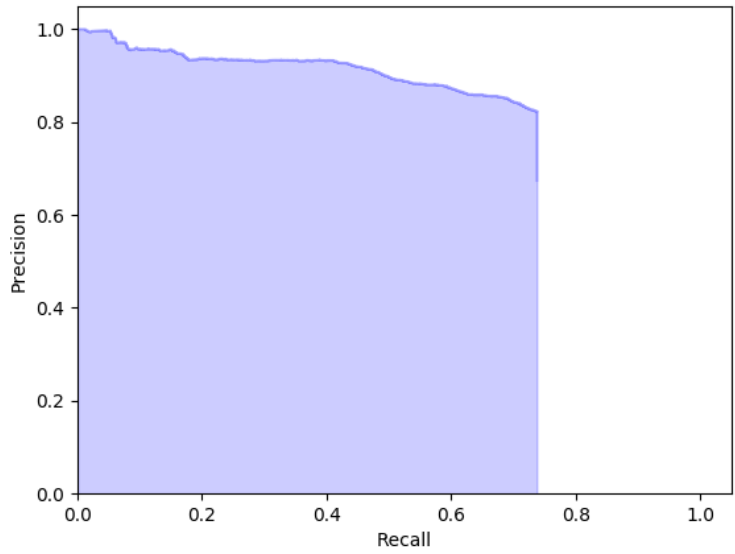

**Figure 3.** Precision-recall curves of the best model (upper: exact match, lower: partial match)